\pdfoutput=1

\documentclass[11pt]{article}

\usepackage[final]{acl}

\usepackage{times}
\usepackage{latexsym}

\usepackage{booktabs}
\usepackage{graphicx}
\usepackage[T1]{fontenc}

\usepackage[utf8]{inputenc}

\usepackage{microtype}

\usepackage{inconsolata}

\usepackage{algorithm}
\usepackage{algpseudocode} 
\usepackage{amsmath}
\usepackage{enumitem}

\usepackage{comment}
\usepackage{macro}
\usepackage{color}
\usepackage[normalem]{ulem}

\title{Halu-NLP at SemEval-2024 Task 6: MetaCheckGPT - A Multi-task Hallucination Detection Using LLM Uncertainty and Meta-models}

\author{Rahul Mehta\thanks{*Equal contribution} \\
 \small IIIT Hyderabad, India \\
 \small\texttt{rahul.mehta@research.iiit.ac.in} \And
 Andrew Hoblitzell\footnotemark[1] \\
 \small Purdue University, USA \\
 \small\texttt{ahoblitz@purdue.edu} \AND
 Jack O'Keefe \\
 \small Northwestern University, USA \\
 \small\texttt{jackokeefe2024@u.northwestern.edu} \And
 Hyeju Jang \\
 \small Indiana University Indianapolis, USA \\
 \small\texttt{hyejuj@iu.edu} \And
 Vasudeva Varma \\
 \small IIIT Hyderabad, India \\
 \small\texttt{vv@iiit.ac.in} \\
 }

\begin{document}
\maketitle

\begin{abstract}
Hallucinations in large language models (LLMs) have recently become a significant problem. A recent effort in this direction is a shared task at Semeval 2024 Task 6, \textbf{SHROOM}, a \textit{ Shared-task on Hallucinations and Related Observable Overgeneration Mistakes \cite{mickus-EtAl:2024:SemEval2024}}. This paper describes our winning solution ranked \textit{1st} and \textit{2nd} in the 2 sub-tasks of model agnostic and model aware tracks respectively. We propose a meta-regressor framework of LLMs for model evaluation and integration that achieves the highest scores on the leader board. We also experiment with various transformer based models and black box methods like ChatGPT, Vectara, and others. In addition, we perform an error analysis comparing GPT4 against our best model which shows the limitations of the former.

\end{abstract}

\section{Introduction}
The recent rapid deployment of large language models (LLMs) has led to a hallucination proliferation which poses a barrier to the reliability and trustworthiness of LLMs \cite{lin2022truthfulqa}. One of the widely agreed upon definition of hallucinations \cite{maynez-etal-2020-faithfulness,xiao2021hallucination} is output text containing information not relevant to the input or a desired output.
Hallucinations should not be thought of as an occasional nuisance, but rather as a systemic issue inherent to these models and their web-sourced training data which can be rife with bias and misinformation. This can directly cause user discontent when these systems are implemented in production or real-world scenarios.

These type of hallucinations have been widely studied in the context of text related tasks like machine translation \cite{dale2022detecting,guerreiro2023hallucinations,guerreiro2023looking}, summarization \cite{huang2023factual,vanderpoel2022mutual} and dialogue generation \cite{shuster-etal-2021-retrieval-augmentation}. Gaps in hallucination detection methods in LLM outputs persist across many such tasks.

Despite some progress in hallucination detection, existing methods may rely upon comparisons to reference texts, overly simplified statistical measures, reliance upon individual models, or annotated datasets which can limit their extensibility. Our approach leverages the uncertainty signals present in a diverse basket of LLMs to detect hallucinations more robustly.

\begin{figure*}
\centering
\includegraphics[scale=0.42]{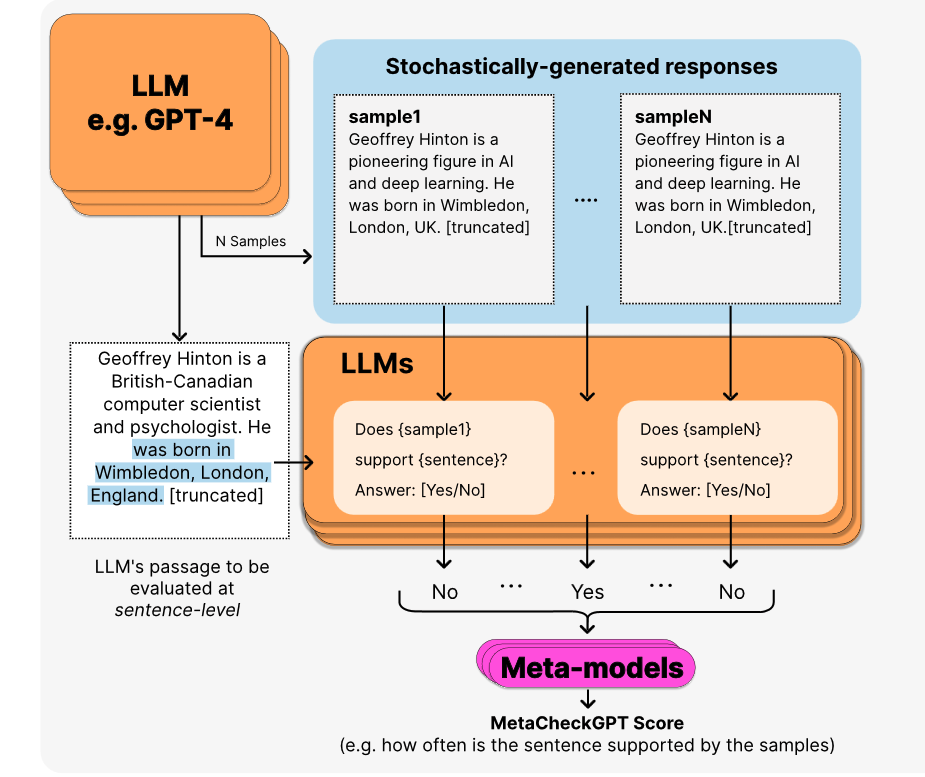}
\caption{MetaCheckGPT: Generated sentences are compared against stochastically generated responses.}
\label{fig:metacheck}
\end{figure*}

In this paper, we present a meta-regressor framework for LLM model selection, evaluation, and integration.\footnote{The code of MetaCheckGPT is available at https://github.com/rahcode7/semeval-shroom} The overall approach is depicted in Figure \ref{fig:metacheck}. For the first step, each LLM-generated sentence is compared against stochastically generated responses with no external database as with SelfCheckGPT \cite{manakul2023selfcheckgpt}. A meta-model that leverages input from a diverse panel of expert evaluators evaluates and integrates the output of multiple iterations of the process.

Our framework focuses on creating a meta-model for identifying hallucinations, with the idea that the meta-model's prediction power is linked to the performance of the underlying base models. This model achieves the highest scores in the SemEval-2024 Task 6 competition across three sub-tasks: Machine translation, Paraphrase generation, and Definition modeling.

\section{Related Work}
In this section, we describe prior work on hallucination detection methods. We will examine two potential streams for hallucination detection: model-aware detection and black-box detection. Model-aware techniques have access to the model's internals, such as weights and logits while black-box methods do not has access to such model internals.

\subsection{Model aware Detection}
These methods require access to model weights and their logits \cite{vanderpoel2022mutual}.
For machine translation task, \citet{guerreiro2023looking} showcased that sequence log-probability performs quite well compared to reference based methods. For article generation task, \citep{varshney2023stitch} uncertainty estimation techniques\cite{azaria2023internal} \cite{Tian2023JustAsk} were used to detect hallucination in ChatGPT. Other methods to detect hallucinations include Retrieval Augmented Generation \cite{ShusterPoffChenKielaWeston2021} and Chain of Verification based techniques (e.g., \cite{LeiLiHuWangYunChingKamal2023}).


\subsection{Black box Detection}
With the prevalence of closed source models, there has been recent work on black-box hallucination detection methods which doesn't require the model inputs, only the generated text.
For example, a recently proposed system SelfCheckGPT \cite{manakul2023selfcheckgpt} utilizes a sampling-based technique based on the idea that sampled responses for hallucinated sentences will contradict each other. 


This model achieves the highest scores across two sub-tasks: Machine translation, Paraphrase generation, and Definition modeling. We perform extensive studies of LLMs like ChatGPT, Mistral, and others to showcase their failure points.

\section{Task Description and Datasets}

In the SHROOM Task-6, the organizers propose a binary classification task to predict a machine generated sentence is a hallucination or not.

The organizers considered 3 types of text generation tasks - Definition Modelling, Machine Translation and Paraphrase Generation. 

\subsection{Task Tracks}
\begin{figure*}
\centering
\includegraphics[scale=0.22]{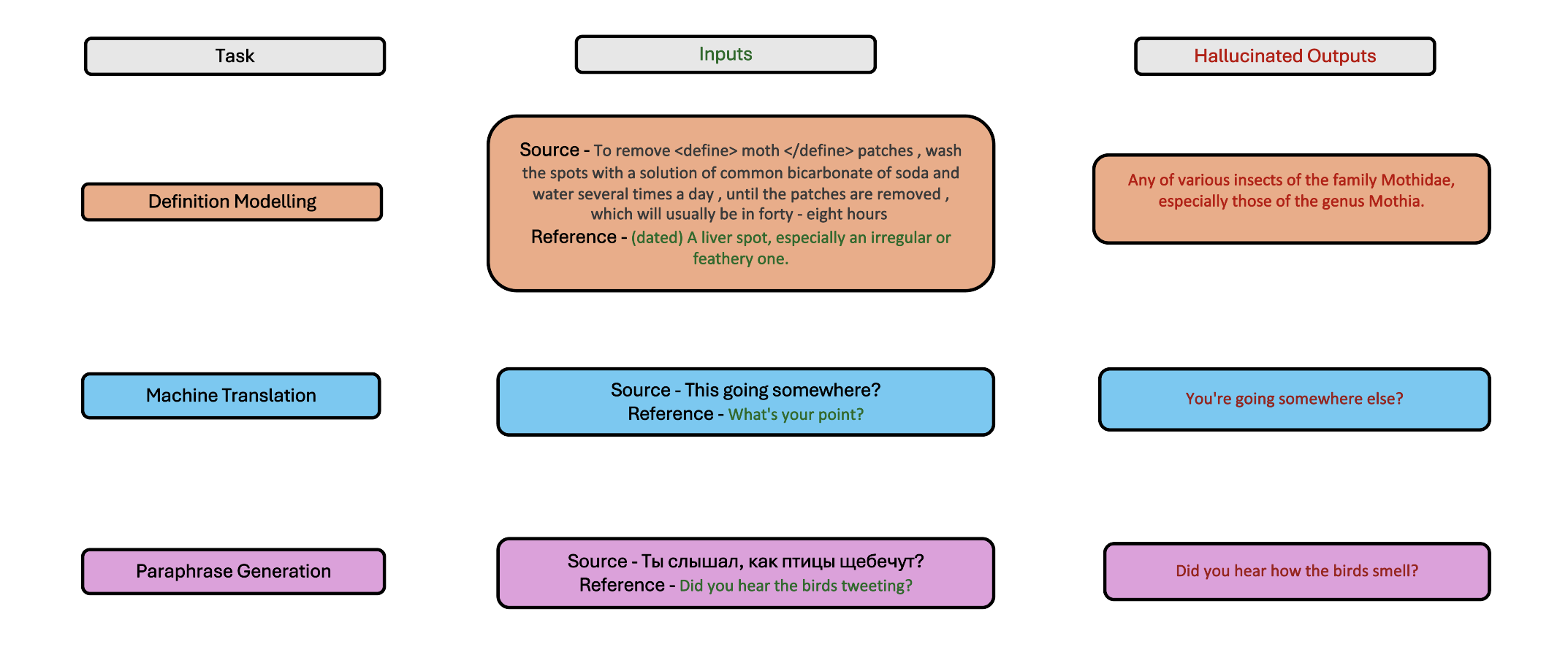}
\caption{Hallucination examples for each task type
}
\label{fig:exam}
\end{figure*}

The shared task was further divided into 2 tracks: \textbf{model agnostic} and \textbf{model aware}. Figure \ref{fig:exam} describes sample examples of hallucinations containing source, reference and output text for each task type.


\subsubsection{Model Agnostic Track}
In this track, only the inputs, references and outputs were provided The details of the model that produced the text was masked from the participants. For data preparation, the SHROOM organizers followed the structure described in \cite{bevilacqua-etal-2020-generationary}.

\begin{table}[ht]
\centering
\begin{tabular}{lccc} 
 \toprule
 & \multicolumn{3}{c}{Model Agnostic Track} \\
 \midrule
 Task & Train & Dev & Test \\
 \midrule
 Definition Modeling & 10000 & 187 & 562 \\
 Machine Translation & 10000 & 187 & 563 \\
 Paraphrase Generation & 10000 & 125 & 375 \\
 \midrule
 Total & 30000 & 499 & 1500 \\
 \bottomrule
\end{tabular}
\caption{Sample counts for the Model Agnostic Track 
}
\label{tab:table1}
\end{table}

\subsubsection{Model Aware Track}
In this track, along with the inputs, references and outputs, the model name and its checkpoints were also provided from which the outputs were generated.

\begin{table}[ht]
\centering
\begin{tabular}{lccc} 
 \toprule
 & \multicolumn{3}{c}{Model Aware Track} \\
 \midrule
 Task & Train & Dev & Test \\
 \midrule
 Definition Modeling & 10000 & 188 & 562 \\
 Machine Translation & 10000 & 188 & 563 \\
 Paraphrase Generation & 10000 & 125 & 375 \\
 \midrule
 Total & 30000 & 501 & 1500 \\
 \bottomrule
\end{tabular}
\caption{Sample Statistics for the Model Aware Track}
\label{tab:table2}
\end{table}

It is worthwhile to note that the organizers chose to share the training which was not labeled and only the development set was labeled.

\section{Our Methodology}
\begin{algorithm}
\caption{Meta-Model Training/Evaluation}
\begin{algorithmic}[1]
\State \textbf{Input:} Base models $M$, Meta-models $N$, Threshold $x$
\State \textbf{Output:} Top performing meta-model
\For{each base model $m$ in $M$}
 \State $score_m \leftarrow$ Evaluate $m$ (MAE)
\EndFor
\State $FilteredMs \leftarrow$ $Models$.filter(MAE $< x$)
\For{each meta-model $n$ in $N$}
 \State Train $n$ with $FilteredMs$
 \State $metaScore_n \leftarrow$ Spearman MAE
\EndFor
\State $TopMeta \leftarrow$ Meta-model in $N$ with lowest Spearman MAE
\end{algorithmic}
\end{algorithm}

Our approach is centered around building a meta-model for hallucination detection, with the hypothesis that the quality of prediction from underlying base models is highly correlated with the meta-model's predictive power. Given a set of base models $M = \{m_1, m_2, ..., m_n\}$ and actual labels $L = \{l_1, l_2, ..., l_n\}$ in the dataset, the Spearman correlation between predicted hallucination scores $H$ and actual labels is given by:

$$
\rho_{s} (H,L) = 1 - \frac {6 \sum d_i^2}{n(n^2-1)}
$$

where $d_i$ is the difference between the ranks of corresponding elements in H and L.

Our overall process was to identify the meta-model that minimized this mean absolute error (MAE) function $\epsilon$, where 
$$
\epsilon = \frac{1}{n} \sum_{i=1}^{n} (Y_i - \hat{Y}_i)
$$
because Spearman correlation was one of the secondary metrics for Task 6 evaluation. Here, $Y_i$ represents the actual Spearman correlation values for hallucination and $\hat{Y}_i$ represents the predicted values. Our overall process is captured in Algorithm 1. 

Algorithms \ref{alg:mr}, \ref{alg:enhanced_mr}, and \ref{alg:mr3} detail how some of our different meta-models were trained. These algorithms follow a unified framework, initiating with the setup of training data and labels, with the ultimate aim of fine-tuning a meta-regressor model. A meta-search cross-validation approach was used to conduct a hyperparameter space for each model's architecture. The process involves iterating over the defined hyperparameter space for each algorithm, fitting the meta-regressor with the training data, and concluding with the identification and preparation of the highest-performing model for deployment. The training process for selecting meta-models is included in the Appendix. RMSE, MAE, and R-squared were used as additional proxies in meta-model evaluation. 

Because this problem was assessed with binary classification accuracy, data was classified based on the Spearman correlation coefficient according to:
$$
\text{Class} = 
\begin{cases}
\text{'Hallucination'}, & 
\rho_s>0.5 \\
\text{'Not Hallucination'}, & \text{otherwise}
\end{cases}
$$
to convert our regression problem into a binary classification task, simplifying the analysis and interpretation of results. Once converted to a classification problem, the primary metric used for evaluation was accuracy. Precision, Recall, F1 Score, and a confusion matrix were used for secondary evaluation.

\section{Experiments \& Results}
\subsection{Experimental set-up}
Training was conducted both on cloud using APIs as well as locally on V100/A100 GPUs for faster processing times. 

We conducted our initial experiments with simpler base models including DeBERTa \cite{he2021deberta}, DistilBERT \cite{sanh2020distilbert}, RoBERTa \cite{liu2019roberta}, LLaMA 2 \cite{touvron2023llama}, and Mixtral of Experts \cite{jiang2024mixtral} among others. Preliminary results indicated an accuracy of 0.5 to 0.6, prompting us to continue our search for more performant base models.

Additional analysis indicated our base models ChatGPT\cite{OpenAIGPT4}, SelfCheckGPT\cite{manakul2023selfcheckgpt} and Vectara\cite{hughes2023cut} showed promising results in initial tests, with accuracy in the range of 0.6 to 0.7. Prompt engineering, self-consistency checks and uncertainty based modeling techniques were used to maximize performance in base models. The training process for more performant meta-models, including random forest and elementary neural ensemble models, can be found in the Appendix.

\subsection{Results}
Classification performance obtained on the training data, which includes an accuracy of 0.8317, precision of 0.7447, recall of 0.875, and an F1 score of 0.8046.

\begin{table}[ht]
\centering
\begin{tabular}{c|c|c|}
\cline{2-3}
& Positive & Negative \\ \hline
\multicolumn{1}{ |c| }{Positive} & TP: 49 & FN: 5 \\ \hline
\multicolumn{1}{ |c| }{Negative} & FP: 12 & TN: 35 \\ \hline
\end{tabular}
\end{table}

Cross-validation and regularization techniques were applied to increase confidence that the performance observed on the training data would be maintained on test data. 

\begin{table}[ht]
 \centering 
 \begin{tabular}{|c|c|c|c|} 
 \hline
 Track & Accuracy & Rho & Rank \\ 
 \hline
 Aware & 80.6 & 0.71 & 1/46 \\
 \hline 
 Agnostic & 84.7 & 0.77 & 2/49 \\
 \hline 
 \end{tabular}
 \caption{Final Modeling results on the test set}
 \label{tab:table3}
\end{table}

\subsection{Discussion}
Our results, as summarized in Table \ref{tab:table3}, demonstrate the effectiveness of meta-regressor models in detecting hallucinations across various text generation tasks. One of the key strengths of the approach is that a diverse set of base models is able to better capture a wide range of features indicative of hallucinations than a single model or knowledge base alone may be able to. High performance metrics underline the promise of combining base models/knowledge bases through meta-learning.

Our approach is not without its limitations. The black-box nature of some base models (e.g. GPT4), limits understanding of the internal mechanisms driving the generation and detection of hallucinations. More detailed limitations of the system and directions for future work are examined in the following section.

\subsection{Limitations}
There are several limitations to the current work. For example, all language models have inherent limitations such as bias and lack of world grounding. Unfortunately, more recent models such as GPT have also started to function as black box systems. The corpus for training data for base language models is predominantly English. The system also would not readily integrate into a production system without additional effort. The system could also benefit from the ability to learn from feedback. All of the base language models may also suffer from potential safety issues like false confidence and over-reliance, etc. 

\section{Conclusion}
Our meta-model strategy represents a step forward in addressing the challenges of mitigating hallucinations and the importance of a nuanced approach to model selection, evaluation, and integration. The work also acknowledges the need for additional research into more transparent, interpretable, and multilingual models, as well as the integration of external knowledge sources and feedback mechanisms to refine and improve hallucination detection methods. In the future, some areas we would like to work on include utilizing additional multilingual datasets, expand the scope of our work to more set of text generation task, and focus more on white box hallucination detection systems.

While the current system was tested on some machine translation tasks, we believe it could benefit from more work on multilingual datasets. The current system could improve by integrating with external knowledge bases via retrieval augmented generation. The system could also be made more usable by distilling its knowledge into a portable fine-tuned model widely available to others. Another area for potential improvement includes integration of human or agent feedback through reinforcement learning.

\section*{Acknowledgements}
We would like to thank the Machine Learning Collective (MLC), a group which promotes collaboration and mentorship opportunities being accessible and free for all, for helping connect us as researchers. We would also especially like to thank Timothee Mickus for being very responsive to questions on the listserv. We would also like to thank Elaine Zosa, Raúl Vázquez, Jörg Tiedemann, Vincent Segonne, Alessandro Raganato, and Marianna Apidianaki for organizing the Shared-task on Hallucinations and Related Observable Overgeneration Mistakes, we felt delighted to have a forum to work with others who have a shared interest in language model capabilities. Thanks to Steven Bethard, Ryan Cotterell, Rui Yanfrom, and many others for their work on the template which we adapted from.

\bibliography{acl_latex}

\newpage
\appendix
\section{Appendix: MR Training Processes}
\label{sec:appendix}

\begin{algorithm}[H]
\caption{MR1 Training: Algorithm 2 outlines the process of training a meta-regressor model with hyperparameters for random forest.}
\label{alg:mr}
\begin{algorithmic}[1]
\Require $X_{train}, y_{train}$ \Comment{Training data and labels}
\Ensure $model_{best}$ \Comment{Optimally tuned model}
\State $MR \gets MetaRegressor()$
\State $H \gets \{n\_estimators \in \{\alpha_1, \ldots, \alpha_N\},$
\State $\quad max\_depth \in \{\beta_1, \ldots, \beta_M\},$
\State $\quad min\_samples\_split \in \{\gamma_1, \ldots, \gamma_L\},$
\State $\quad min\_samples\_leaf \in \{\delta_1, \ldots, \delta_K\},$
\State $\quad max\_features \in \{\text{'auto'}, \text{'sqrt'}\},$
\State $\quad bootstrap \in \{\text{True}, \text{False}\}\}$
\State $MetaCV = MetaSearchCV(MR, H, cv)$
\State $MetaCV.fit(X_{train}, y_{train})$
\State $params_{best} = MetaCV.best_params$
\State $model_{best} = MetaRegressor(params_{best})$
\State $model_{best}.fit(X_{train}, y_{train})$
\end{algorithmic}
\end{algorithm}

\begin{algorithm}[H]
\caption{MR2 Training: Algorithm 3 outlines the process of training a meta-regressor model with hyperparameters for gradient boosted trees.}
\label{alg:enhanced_mr}
\begin{algorithmic}[1]
\Require $X_{train}, y_{train}$ \Comment{Training data and labels}
\Ensure $model_{best}$ \Comment{Optimally tuned model}
\State $MR \gets MetaRegressor()$
\State $H \gets \{ n\_estimators \in \eta_1, \ldots, \eta_n,$
\State $\quad learning\_rate \in {\theta_1, \ldots, \theta_n},$
\State $\quad max\_depth \in {\iota_1, \ldots, \iota_n},$
\State $\quad min\_child\_weight \in {\kappa_1, \ldots, \kappa_n},$
\State $\quad gamma \in {\lambda_1, \ldots, \lambda_n},$
\State $\quad subsample \in {\mu_1, \ldots, \mu_n},$
\State $\quad colsample\_bytree \in {\nu_1, \ldots, \nu_n},$
\State $\quad reg\_alpha \in {\xi_1, \ldots, \xi_n},$
\State $\quad reg\_lambda \in \zeta_1, \ldots, \zeta_n\}$
\State $MetaCV = MetaSearchCV(MR, H, cv)$
\State $MetaCV.fit(X_{train}, y_{train})$
\State $params_{best} = MetaCV.best_params$
\State $model_{best} = MetaRegressor(params_{best})$
\State $model_{best}.fit(X_{train}, y_{train})$
\end{algorithmic}
\end{algorithm}

\begin{algorithm}[H]
\caption{MR3 Training: Algorithm 4 the training procedure for a meta-regressor model designed for an elementary neural ensemble model.}
\label{alg:mr3}
\begin{algorithmic}[1]
\Require $X_{train}, y_{train}$ \Comment{Training data and labels}
\Ensure $model_{best}$ \Comment{Optimally tuned model}
\State $MR \gets MetaRegressor()$
\State $H \gets \{ num\_layers \in {\eta_1, \ldots, \eta_n},$
\State $\quad \textbf{For each layer } i \textbf{ in } {1, \ldots, num\_layers}:$
\State $\quad \quad units\_i \in {\delta_1, \ldots, \delta_n},$
\State $\quad \quad activation\_i \in {\zeta_1, \ldots, \zeta_n},$
\State $\quad \quad l2\_reg \in {\iota_1, \ldots, \iota_n},$ 
\State $\quad dropout \in {\gamma_1, \ldots, \gamma_n}$
\State $\quad learning\_rate \in {\theta_1, \ldots, \theta_n},$ \}
\State $MetaCV = MetaSearchCV(MR, H, cv)$
\State $MetaCV.fit(X_{train}, y_{train})$
\State $params_{best} = MetaCV.best_params$
\State $model_{best} = MetaRegressor(params_{best})$
\State $model_{best}.fit(X_{train}, y_{train})$
\end{algorithmic}
\end{algorithm}

\end{document}